%% file: arxiv.tex
\documentclass[twocolumn]{article}

\usepackage[utf8]{inputenc}
\usepackage[english]{babel}

\usepackage{hyperref,verbatim,graphicx,amsmath,amssymb,dsfont,amsthm}
\usepackage{graphicx,subcaption}
\usepackage{stackengine}
\usepackage{bm}
\usepackage{bbold}
\usepackage{dsfont}
\usepackage{booktabs}
\usepackage{tikz}
\usetikzlibrary{shapes,arrows}
\usepackage{nicefrac}

\usepackage{tikz}
\usetikzlibrary{spy}
\newcommand{\spied}[2]{\begin{tikzpicture}[every node/.style={inner sep=0,outer sep=0},spy using  outlines={white,magnification=8,size=1.0cm, connect spies}]
\node {\pgfimage[width=\textwidth]{#2}} ;
\spy on #1 in node at (current bounding box.south west) [anchor=south west];
\end{tikzpicture}}

\title{Parallax estimation for push-frame satellite imagery: application to super-resolution and 3D surface modeling from Skysat products}

\author{Jérémy Anger$^{1,2}$ \qquad Thibaud Ehret$^1$ \qquad Gabriele Facciolo$^1$}
\date{$^1$Université Paris-Saclay, CNRS, ENS Paris-Saclay, Centre Borelli, France
\\ $^2$Kayrros SAS}

\begin{document}

\maketitle

\begin{abstract}
Recent constellations of satellites, including the Skysat constellation, are able to acquire burst of images.
This new acquisition mode allows for modern image restoration techniques, including multi-frame super-resolution.
As the satellite moves during the acquisition of the burst, elevation changes in the scene translate into noticeable parallax.
This parallax hinders the results of the restoration.
To cope with this issue, we propose a novel parallax estimation method.
The method is composed of a linear Plane+Parallax decomposition of the apparent motion and a multi-frame optical flow algorithm that exploits all frames simultaneously.
Using SkySat L1A images, we show that the estimated per-pixel displacements are important for applying multi-frame super-resolution on scenes containing elevation changes and that can also be used to estimate a coarse 3D surface model.
\end{abstract}

\section{Introduction}

Satellites play a big role in the observation of the Earth: from environmental monitoring to industry monitoring.
Many Earth monitoring applications require a good ground resolution. For example, fine detection and analysis of human activity requires a resolution in the range of 30~cm to 1~m~/~pixel~\cite{murthy2014Skysat}.

Earth observation missions were historically owned by national organisations, constructing high-cost long-term satellites. Starting in the late 90's, a similar model was also adopted by actors from the private sector  (\textit{e.g.} IKONOS, EROS, QuickBird, WorldView). But in recent years, some companies have started to offer low-cost imagery thanks to new satellite designs.
The current trend is to launch many smaller satellites to lower orbits and  with a shorter lifespan, providing a wider coverage at lower cost.
For example, this strategy allows Planet to provide a daily revisit time on some products. However, low-cost satellites means that the quality of each individual image is lower, with higher noise or worse GSD for example. 
This means that instead of trying to obtain a 50~cm GSD from the physical design of the satellite, such resolution has to be reached using computational photography techniques such as multi-frame super-resolution.

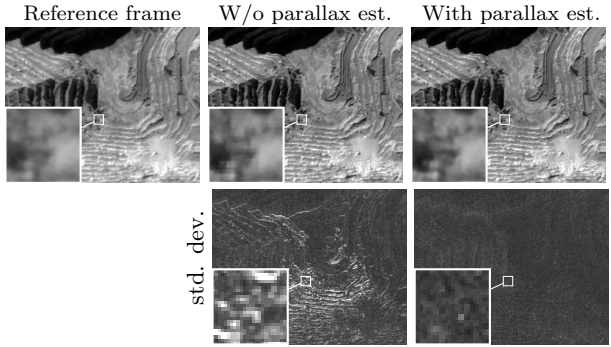
\begin{figure}
    \centering
    \def\b{mine}
    \def\pos{(-0.05,-0.19)}
    \centering\rotatebox{0}{\footnotesize{~~~Reference frame~~~~~W/o parallax est.~~~~With parallax est.~~~~~~}}

    \foreach \i in {l1a,noparallax,multiflow2}{%
        \begin{subfigure}{0.32\linewidth}
            \spied{\pos}{scripts/results_cropped/\b_\i_sr.png}
        \end{subfigure}
    }\\[0.1em]
    \begin{subfigure}{0.28\linewidth}
        \begin{tikzpicture}
        \end{tikzpicture}
    \end{subfigure}
    \centering~\rotatebox{90}{\small\hspace{-1.25em}std. dev.}
    \foreach \i in {noparallax_std,multiflow2_std}{%
        \begin{subfigure}{0.32\linewidth}
            \spied{\pos}{scripts/results_cropped/\b_\i_sr.png}
        \end{subfigure}
    }%
    \caption{Accurate parallax estimation for super-resolution allows to restore finer details. Top row: reference low-resolution frame, SR without parallax estimation, SR with parallax estimation using the proposed method. Bottom row: standard deviation of the frames aligned with the estimated displacements.}\label{fig:mine-sr}
\end{figure}

The SkySat constellation~\cite{murthy2014Skysat} from Planet follows this trend towards small but high-resolution satellites. SkySats contain a full-frame sensor  able to 
capture 40 frames per second and is operated in a \emph{push-frame} mode with significant overlap between the frames.
This means that the same point on the ground is seen in several consecutive images, which are combined with computational photography techniques. 
Furthermore, thanks to the design of its optical system, the images are aliased, providing the opportunity to increase the resolution by means of multi-frame super-resolution.

A simple strategy to multi-frame super-resolution consists in aligning the frames using a rigid model (assuming that the observed scene is flat) and then fusing them~\cite{anger2020}.  
However, because the frames are acquired from slightly different viewpoints, when the scene contains large elevation changes the parallax effect due to surface relief becomes noticeable and can hinder the reconstruction.
Indeed, as noted in~\cite{delon2007small}, the parallax $\delta \varepsilon$ is directly proportional to the  elevation changes $\delta z$, inversely proportional to the altitude $h$ of the satellite, and proportional to the distance between the viewpoints $b$ (baseline)
\begin{equation}
    \delta \varepsilon \propto \frac{b}{h} \delta z.
    \label{eq:boverh}
\end{equation}
Since the overlapping frames are acquired with increasingly larger baselines the parallax will introduce non rigid displacements of the samples proportional to their elevation.

In order to avoid fusion artefacts, this parallax has to be estimated and taken into account in the super-resolution. 
In a calibrated system (with perfectly known camera models) plus knowledge of the scene topography this displacement could be easily deduced. However, since camera models are subject to pointing errors a solution for non-calibrated systems is needed.

The problem of estimating the 3D geometry of a scene from a set of views from different angles is classic in computer vision. It usually involves first the estimation of the camera parameters via bundle adjustment~\cite{triggs2000plane}, then matching across the views and triangulating the matches~\cite{s2p}. 
However, Equation~\eqref{eq:boverh} implies that the projective camera model can be assumed affine~\cite{de2014stereo}. In addition, because of the push-frame acquisition the correspondences between frames can be reliably estimated using optical flow methods.

\textbf{Contributions.}
In this paper we propose to estimate the parallax from a burst of non-calibrated frames using a linear \emph{Plane+Parallax} formulation~\cite{Irani2002}, where the correspondences between frames are provided by an optical flow algorithm.
The frames are first aligned with respect to a common virtual plane, from which the parallax between all frames can then be estimated using a multi-frame optical flow algorithm.
We show that the estimated parallax allows to improve super-resolution results on scenes with large elevation changes and moving objects as shown in Figure~\ref{fig:mine-sr}. In addition, we show that the estimated parallax can provide coarse 3D information, even if the frames are not acquired in a stereo setting.

\section{Related works}

The video product of the SkySat constellation was used in~\cite{wan20163d} to estimate a depth map.
The authors computed several low baseline disparity maps using phase correlation, which were then warped and stacked to improve the SNR of the final disparity map. The warp was computed by matching the images to a reference one. 
A similar  study  was also conducted in~\cite{dangelo2014evaluation} with the objective of producing an elevation model also using a SkySat video product. 
 SkySat video products require a satellite maneuver to keep the object of interest in the frame during the video, resulting in thousands of frames available.
In contrast, our method exploits the redundancy in the L1A SkySat product, which consists of a few dozen frames for a given scene.
This implies that the maximum available baseline  is much smaller, resulting in a more ill-posed estimation.

In~\cite{Irani2002} it is proposed a method for estimating Plane+Parallax motion from multiple views of a 3D scene. The method alternates between computation of the parallax motion  common to all the frames and the camera epipoles. For that they use a photometric constraint between the reference and all the other images.  Our method relies on a similar joint multi-frame formulation, but since we assume affine cameras there is no need to alternate with the epipole estimation and the parallax can be directly recovered.

\section{Method}

The inputs of the proposed method are a set of satellite images acquired in a push-frame burst (as possible by the SkySat constellation~\cite{murthy2014Skysat}) and pre-registered using a translation. The objective is to estimate the parallax in the scene.

In~\cite{anger2020} an affine  transform is estimated between each frame and the reference one.
This registration is not sufficient when the scene contains large elevation changes.
To account for these, we propose a refined model based on one  affine transform per frame noted $A_i$ and a common dense optical flow $d$ to model the geometry of the scene.

\subsection{Plane stabilization}

In this section we study how to register the frames in presence of parallax.
The frames are assumed to be  taken at uniform intervals and along a rectilinear trajectory and  registered up to a translation.
Since the satellite is far from the scene we can assume that parallax induces a disparity translation $d$ proportional to the height of the ground~\cite{delon2007small} and proportional to the  baseline between the frames (i.e. proportional to the frame number)
\begin{align}\label{eq:plane-stab-eq1}
    v_0(x) = v_i(x + i \cdot d(x)).
\end{align}

Equation~\ref{eq:plane-stab-eq1} corresponds to the case of a perfectly rectified scene, where the same zero-parallax plane is always registered in the sequence $v_i$.
However, due to pointing errors and the limitations of the registration up to a translation, the actual model might include an additional homography $H_i$, which we can safely  approximate by an affine transform $A_i$ because of the distance from the satellite to the scene~\cite{de2014stereo}:
\begin{align}
    v_0(x) = v_i( A_i(x + i\cdot d(x)) ).
    \label{eq:realmodel}
\end{align}
The transforms $A_i$ and the vector field $d$ are unknown, and must be estimated.
Note that in~\eqref{eq:realmodel} the disparity $d$ is composed with the affinities $A_i$. However, since the $A_i$ are close to translations (with Jacobian $J_i$ close to the identity) we can safely use the following linear approximation of the previous model  
\begin{align}
    v_0(x) = v_i( A_i(x) +  J_i (i\cdot d(x)) ),
\end{align}
where we assume that $J_i = \mathds{1}$. 
The advantage of this formulation is that it leads to a linear system  of equations on  the parameters of $A_i$ and $d$. 
When $A_i\neq \mathds{1}$ the linearization could still be applied but an rough estimate of $A_i$ should be provided (for instance by aligning the frames~\cite{ica}) from which the Jacobians can be extracted. 
We now detail how our linear  system is built and solved. 

First, the pairwise optical flows $f_i$ are computed between the frames $u_i$ and frame $u_0$ and thus verify the  relations
\begin{align}
    v_0(x) = v_i( f_i (x) ).
\end{align}
The flows $f_i$ provide correspondences between the frames, hence we want to factorize them as
\begin{equation}
   f_i(x) =  A_i (x) + i \cdot d(x).
\end{equation}

The set of equations associated to all the flows form an over-determined linear system that can be solved by least squares minimization. We solve for all the $A_i$ and a single $d$ using the Conjugate Gradient method.
In order to speed-up this optimization, the linear system can be sub-sampled (e.g.~by a factor 4, thus considering only
\nicefrac{1}{16}
equations).

At this point the flow $d$ shows some part of the parallax, but is not very precise due to the noisy nature of each individual flows $f_i$.
However, the affine transformations $A_i$ allow to align the frames to a common plane, which is necessary for the next step: parallax estimation.

\subsection{Parallax estimation}

As a result of the previous step, the affine transforms $A_i$ can be used to align the frames $v_i$ on a common plane.
For the parallax estimation step, we propose a multi-frame optical flow method, allowing to fully exploit all the frames to produce a single disparity map.

The proposed method is an extension of the robust optical flow method~\cite{robustof} to multiple frames assuming a constant motion model.
The flow is computed using all the images at the same time, contrary to a regular optical flow that is computed between pairs of frames.
As before, we consider a set of $N$ frames $\{v_{-N/2}, \dots, v_0, \dots, v_{N/2}\}$ such that $v_0$ corresponds to the reference frame and that the frames are now aligned on a common plane using the transforms $A_i$.
Given that the movement computed corresponds only to the parallax effect, we assume that the optical flow between the frame $0$ and the $i$th frame, $w_{0,i}$ is linearly proportional to the optical flow from $0$ to $1$ such that
$w_{0,i} = i \cdot w_{0,1}$.
This indicates that a single optical flow capturing the parallax effect is able to align each frame to the reference.

Let $d$ be the optical flow to be estimated.
The energy of the robust optical flow method~\cite{robustof} for a single pair $(v_0, v_i)$ is
\begin{gather}
\begin{aligned}
    E_i(d) &= \sum_x \Psi(v_i(A_i x+i \cdot d) - v_0(A_0 x))
    \\&+ \gamma \sum_x \Psi(\nabla v_i(A_i x+i\cdot d) - \nabla v_0(A_0 x))
    \\&+ \alpha \sum_x \Psi(\sqrt(|\nabla d_x|^2 + |\nabla d_y|^2))
\end{aligned}
\end{gather}
with
    $\Psi(x) = \sqrt{x^2 + \epsilon^2}$
and $d_x$ (respectively $d_y$) corresponds to the $x$ (respectively $y$) component of the optical flow.
Finding $d$ that fits the best all frames corresponds then to minimizing the following energy
\begin{equation}
    E(d) = \frac{1}{N}\sum_{i=-N/2}^{N/2} E_i(d),
\end{equation}
which is optimized using \textit{successive over-relaxation}~\cite{robustof}.

\section{Experiments}

In this section we validate the proposed method on two applications: multi-frame super-resolution and 3D surface modeling.
We show that our parallax estimation method allows for precise displacement estimation between the input full-frame images.

\subsection{Application to super-resolution}
We first demonstrate the proposed method on a multi-frame super-resolution from SkySat burst. Multi-frame super-resolution usually relies on a registration step, which has to be as precise as possible.
For this reason, it is convenient to estimate global parametrized transform (e.g.~affine transform as in~\cite{anger2020}).
However, in the presence of  abrupt elevation changes in the scene, the change of viewpoint of the satellite {between} frames induces a parallax motion. As we have seen, such parallax motion must be estimated locally.

We compare the multi-frame super-resolution method of~\cite{anger2020} without and with our parallax estimation method.
Figure~\ref{fig:mine-sr} shows the result of the two methods on images of an open pit mine near Sahuarita AZ.
We observe that the rigid registration exhibits artefacts due to the elevation change in the scene, whereas the proposed registration is able to retrieve fine details.
Furthermore, this figure also shows the temporal standard deviation of the aligned stack of images (interpolated with splines of order 5).
As the rigid registration is not able to register local structures, the temporal standard deviation is high around such structures. Instead, the proposed method successfully minimizes this measure.
For a quantitative comparison, the average of the temporal standard deviation are reported in Table~\ref{tbl:std}.
We observe that the standard deviation is lower when using a low optical flow regularization $\alpha$, which is important of the context of super-resolution.

\begin{table}
    \caption{Comparison of the average standard deviation of the aligned stacks on three locations.}
    \centering
    \small{%
    \begin{tabular}{lccc}%
        \toprule
        & & \multicolumn{2}{c}{Proposed} \\
        & Rigid transform & $\alpha=10$ & $\alpha=60$ \\
        \midrule
        Sahuarita  & \input{scripts/stds/mine_noparallax.tex}    & \bf\input{scripts/stds/mine_multiflow2.tex}    & \input{scripts/stds/mine_multiflow3.tex}    \\
        Morenci    & \input{scripts/stds/morenci_noparallax.tex} & \bf\input{scripts/stds/morenci_multiflow2.tex} & \input{scripts/stds/morenci_multiflow3.tex} \\
        Chicago    & \input{scripts/stds/chicago_noparallax.tex} & \bf\input{scripts/stds/chicago_multiflow2.tex} & \input{scripts/stds/chicago_multiflow3.tex} \\
        \bottomrule
    \end{tabular}}\label{tbl:std}
\end{table}

\begin{figure}
    \centering
    \def\b{chicago}
    \centering\rotatebox{0}{\footnotesize{~~~Reference frame~~~~~W/o parallax est.~~~~With parallax est.~~~~~}}

    \foreach \i in {l1a,noparallax,multiflow2}{%
        \includegraphics[width=0.32\linewidth]{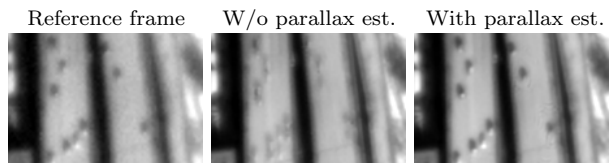}
    }%
    \caption{Parallax estimation allows to register moving objects and remove ghosting artifacts.}\label{fig:chicago-sr-movingobjects}
\end{figure}

\medskip\noindent\textbf{Moving objects.}
Objects that have a constant speed are also captured by the multi-frame optical flow.
In this case, the flow can be used to compute the speed of the objects~\cite{dangelo2014evaluation}, as well as to provide accurate displacements for the super-resolution  algorithm.
Figure~\ref{fig:chicago-sr-movingobjects} shows two super-resolution results, with and without parallax handling ($\alpha=10$).
When fusing images  only using global transforms, the moving objects are accumulated at incorrect positions, resulting in ghosting artifacts.

\subsection{Application to 3D surface modeling}
Per-pixel registration using plane stabilization and multi-frame optical flow provides accurate displacement for each pixel. This displacement is proportional to the elevation of the scene (up-to the choice of the plane by the stabilization), and thus gives a coarse estimation of the elevation model.
In order to  obtain a  smoother result, we set a higher regularization weight of $\alpha=60$ when estimating the multi-frame optical flow for the application of coarse 3D modeling.

\begin{figure}
    \centering
    \def\b{morenci}
    \foreach \i in {multiflow,multiD3,s2p}{%
        \includegraphics[width=0.32\linewidth]{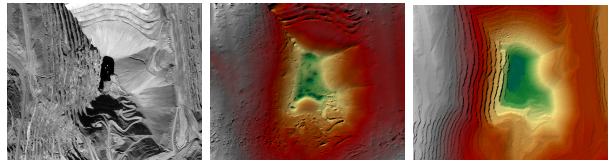}
    }%
    \caption{Qualitative comparison between the result of our coarse 3D modeling and the DSM result of S2P~\cite{s2p}.}\label{fig:morenci}
\end{figure}

Figure~\ref{fig:morenci} shows the resulting parallax estimation on a large open-pit mine near Morenci AZ.
This result is obtained from the multi-frame optical flow estimated using 35 consecutive L1A frames.
We compare our result with a DSM estimated using S2P~\cite{s2p} from the L1B frames acquired in tri-stereo configuration.
While the proposed reconstruction is very coarse, we can identify the main 3D features. 

\section{Conclusion}

We proposed a novel parallax estimation method from SkySat images.
The method is able to estimate precise per-pixel displacements.
When used along a multi-frame super-resolution algorithm,
the proposed method allows to recover more details and contain less artefacts.
Moreover, we showed that parallax estimation from a single burst can be used as a coarse alternative to 3D reconstruction from stereo pairs.

Using images with a small baseline to reconstruct the 3D as we propose is a very ill-posed problem, and our method could probably benefit from more advanced optical flow priors. Future works will focus on reducing the computational of the method, for instance using a faster optical flow implementation. 

\section*{Acknowledgements}
Work partly financed by IDEX Paris-Saclay IDI 2016, ANR-11-IDEX-0003-02, Office  of Naval research grant N00014-17-1-2552, DGA Astrid project  \mbox{\guillemotleft~filmer la Terre~\guillemotright} n\textsuperscript{o}ANR-17-ASTR-0013-01, MENRT. This work was  performed using HPC resources from GENCI–IDRIS (grant 2020-AD011011795). We thank Planet for providing the L1A SkySat images.

\bibliographystyle{plain}
\bibliography{arxiv}

\end{document}

%% file: scripts/stds/mine_noparallax.tex
 0.41795

%% file: scripts/stds/mine_multiflow2.tex
 0.34511

%% file: scripts/stds/mine_multiflow3.tex
 0.34638

%% file: scripts/stds/morenci_noparallax.tex
 0.45282

%% file: scripts/stds/morenci_multiflow2.tex
 0.42532

%% file: scripts/stds/morenci_multiflow3.tex
 0.43164

%% file: scripts/stds/chicago_noparallax.tex
 0.52539

%% file: scripts/stds/chicago_multiflow2.tex
 0.45405

%% file: scripts/stds/chicago_multiflow3.tex
 0.47601